\documentclass[10pt,a4paper]{article}

\usepackage{natbib}
\pdfoutput=1

\usepackage{graphicx}

\usepackage{caption}

\usepackage{subcaption}

\usepackage{authblk}

\title{Recurrent Neural Networks for Time Series Forecasting}
\author{G\'abor Petneh\'azi\thanks{gabor.petnehazi@science.unideb.hu}}
\affil{Doctoral School of Mathematical and Computational Sciences\\
University of Debrecen}
\date{}

\begin{document}
\maketitle

\begin{abstract}
Time series forecasting is difficult. It is difficult even for recurrent neural networks with their inherent ability to learn sequentiality. This article presents a recurrent neural network based time series forecasting framework covering feature engineering, feature importances, point and interval predictions, and forecast evaluation. The description of the method is followed by an empirical study using both LSTM and GRU networks.
\end{abstract}

\section{Introduction}
Recurrent neural networks are well suited to supervised learning problems where the dataset has a sequential nature. Time series forecasting should not be an exception.\\
RNNs are essentially neural networks with memory. They can remember things from the past, which is obviously useful for predicting time-dependent targets. Yet, applying them to time series forecasting is not a trivial task. The aim of this article is to review some practices that can help RNNs deliver useful forecasts.
\section{Neural networks for forecasting}
The objective of supervised learning is to predict something from data. A training set of an output (target) and some input variables is fed to an algorithm that learns to predict the target values. The output may be categorical (classification) or continuous (regression). The task of the algorithm is to deliver high quality predictions, all by itself, extracting the required knowledge solely from the available data.\\
Neural networks are a popular framework for supervised learning --- a network-like system of weighted summations and differentiable functions that can learn astoundingly complex things. Usually, variants of gradient descent together with backpropagation (chain rule) are used to find the optimal values of the network weights. This is all simple and intuitive, yet the resulting networks are usually difficult for people to understand. There are so many weights and connections that we just wonder how the system produced the results. We don't understand neural networks, but we like them. The reason for their ever high popularity is simple: they are good. They can learn arbitrarily complex functions, and they often provide excellent predictions for pretty difficult machine learning problems.\\
NNs are widely used in machine learning, time series prediction is just one example application. \citet{werbos1974beyond} and \citet{werbos1988generalization} made a pioneering work in the field of neural networks by developing a general formulation of backpropagation. Werbos applied the method to forecasting, and compared it to traditional forecasting methods. \citet{tang1991time} also made a neural networks vs. Box-Jenkins comparison and found that NNs outperform the Box-Jenkins model for series with short memory. For series with long memory, both methods produced similar results. \citet{faraway1998time} compared neural networks with Box-Jenkins and Holt-Winters methods for forecasting and found that the design of network architecture and the choice of input variables require great care, and so applying neural networks in black box mode is not a good idea. They also found that increasing the number of hidden nodes may deteriorate out-of-sample performance.\\
\citet{zhang2005neural} found that neural networks are not able to capture seasonality by default, and deseasonalization and detrending can help their forecasting performance. According to \citet{balkin2000automatic}, differencing is unnecessary for neural network based forecasting, but a log transformation may be beneficial.\\
\citet{zhang1998forecasting} gives a detailed review of neural networks for forecasting. \citet{gamboa2017deep} provides a more recent review of the applications of deep learning to time series data.
\section{Recurrent neural networks}
RNNs are neural networks for sequential data --- hereby we apply them to time series. The main idea behind recurrent neural networks is using not only the input data, but also the previous outputs for making the current prediction. This idea makes a lot sense --- we could build neural networks passing values forward in time. However, such simple solutions usually do not work as expected. They are hard to train and they are forgetful. Rather, we need to have a system with some kind of memory.\\
There are two popular and efficient RNN models that work really well: long short-term memory and gated recurrent unit.
\subsection{LSTM}
Long short-term memory (\citet{hochreiter1997long}) is a gated memory unit for neural networks. It has 3 gates that manage the contents of the memory. These gates are simple logistic functions of weighted sums, where the weights might be learnt by backpropagation. It means that, even though it seems a bit complicated, the LSTM perfectly fits into the neural network and its training process. It can learn what it needs to learn, remember what it needs to remember, and recall what it needs to recall, without  any special training or optimization. The input gate (\ref{eq:lstm_inputgate}) and the forget gate (\ref{eq:lstm_forgetgate}) manage the cell state (\ref{eq:lstm_cellstate}), which is the long-term memory. The output gate (\ref{eq:lstm_outputgate}) produces the output vector or hidden state (\ref{eq:lstm_hiddenstate}), which is the memory focused for use. This memory system enables the network to remember for a long time, which was badly missing from vanilla recurrent neural networks.
%lstm
\begin{equation} \label{eq:lstm_inputgate}
i_{t} = sigmoid \left ( W_{i}x_{t} + U_{i}h_{t-1} + b_{i} \right )
\end{equation}
\begin{equation} \label{eq:lstm_forgetgate}
f_{t} = sigmoid \left ( W_{f}x_{t} + U_{f}h_{t-1} + b_{f} \right ) 
\end{equation}
\begin{equation} \label{eq:lstm_outputgate}
o_{t} = sigmoid \left ( W_{o}x_{t} + U_{o}h_{t-1} + b_{o} \right )
\end{equation}
\begin{equation} \label{eq:lstm_cellstate}
c_{t} = f_{t} \odot c_{t-1} + i_{t} \odot tanh \left ( W_{c}x_{t} + U_{c}h_{t-1} + b_{c} \right )
\end{equation}
\begin{equation} \label{eq:lstm_hiddenstate}
h_{t} = o_{t} \odot tanh \left ( c_{t} \right )
\end{equation}
\subsection{GRU}
Gated recurrent unit (\citet{cho2014learning}) is essentially a simplified LSTM. It has the exact same role in the network. The main difference is in the number of gates and weights --- GRU is somewhat simpler. It has 2 gates. Since it does not have an output gate, there is no control over the memory content. The update gate (\ref{eq:gru_updategate}) controls the information flow from the previous activation, and the addition of new information as well (\ref{eq:gru_outputvector}), while the reset gate (\ref{eq:gru_resetgate}) is inserted into the candidate activation. Overall, it is pretty similar to LSTM. From these differences alone, it is hard to tell, which one is the better choice for a given problem. For a comparison, see \citet{chung2014empirical}.
%gru
\begin{equation} \label{eq:gru_updategate}
z_{t} = sigmoid \left ( W_{z}x_{t} + U_{z}h_{t-1} + b_{z} \right )
\end{equation}
\begin{equation} \label{eq:gru_resetgate}
r_{t} = sigmoid \left ( W_{r}x_{t} + U_{r}h_{t-1} + b_{r} \right ) 
\end{equation}
\begin{equation} \label{eq:gru_outputvector}
h_{t} = z_{t} \odot h_{t-1} + (1 - z_{t}) \odot tanh \left ( W_{h}x_{t} + U_{h}(r_{t} \odot h_{t-1}) + b_{h} \right )
\end{equation}
\subsection{Recurrent neural networks for forecasting}
Though it is probably not their primary application, LSTM and GRU networks are often used for time series forecasting. \citet{gers2002learning} used LSTMs with peephole connections to learn temporal distances. \citet{malhotra2015long} used stacked LSTM networks to detect anomalies in time series. \citet{guo2016robust} proposed an adaptive gradient learning method for RNNs that enables them to make robust predictions for time series with outliers and change points. \citet{hsu2017time} incorporated autoencoder into LSTM to improve its forecasting performance. \citet{cinar2017time} proposed an extended attention mechanism to capture periods and model missing values in time series. \citet{bandara2017forecasting} used LSTMs on groups of similar time series identified by clustering techniques. \citet{laptev2017time} applied RNNs to special event forecasting and found that neural networks might be a better choice than classical time series methods when the number, the length and the correlation of the time series are high. \citet{che2018recurrent} built a GRU-based model with a decay mechanism to capture informative missingness in multivariate time series.\\
Several attempts have been made on better understanding RNNs. \citet{karpathy2015visualizing} explored the source of recurrent neural networks' performance with performance comparisons and error analysis on character level language models. \citet{van2017visualizing} used different datasets to visualize the operations of LSTMs. \citet{greff2017lstm} compared the performance of several LSTM variants, and found that the forget gate and the output activation function are the most important elements of an LSTM block. \citet{chang2017dropout} proposed a feature ranking method using variational dropout (\citet{hinton2012improving}).\\
Some studies tried to find ways to measure the uncertainty associated with the time series forecasts of recurrent neural networks. \citet{zhu2017deep} made uncertainty estimates using Monte Carlo dropout and an encoder-decoder framework of LSTM units. \citet{caley2017deep} constructed prediction intervals for convolutional LSTMs using bootstrapping.\\
Here we are going to explore different aspects of RNN-based time series forecasting, and introduce an end-to-end framework for producing meaningful forecasts.
\section{Features}
\subsection{Feature engineering}
LSTM and GRU networks can learn and memorize the characteristics of time series. It is not so easy though, especially when we only have a short series of values to learn from. Smart feature engineering can help.\\
There are very few things whose future values we certainly know. Time is one such thing --- we always know how it passes. Therefore, we can use it to make forecasts, even for multiple steps ahead into the future, without increasing uncertainty. All we have to do is extracting useful features that our algorithm can easily interpret.\\
Time series components, such as trend or seasonality can be encoded into input variables, just like any deterministic event or condition. Time-shifted values of the target variable might also be useful predictors.\\
Features are usually normalized before being fed to the neural network. It is beneficial for the training process. Two popular choices for rescaling the variables are the minmax scaler (\ref{eq:minmax_scaler}) and the standard scaler (\ref{eq:standard_scaler}).
\begin{equation} \label{eq:minmax_scaler}
\tilde{x} = \frac{x - min(x)}{max(x) - min(x)}
\end{equation}
\begin{equation} \label{eq:standard_scaler}
\tilde{x} = \frac{x - mean(x)}{sd(x)}
\end{equation}
\subsubsection{Lags}
Lagging means going some steps back in time. To predict the future, the past is our best resource --- it is not surprising that lagged values of the target variable are quite often used as inputs for forecasting. Lags of any number of time steps might be used. The only drawback of using lagged variables is that we lose the first observations --- those whose shifted values are unknown. This might be a matter when the time series is short.
\subsubsection{Trend}
Trend can be grabbed by features indicating the passage of time. A single variable of equidistant increasing numbers might be enough for that.
\subsubsection{Seasonality}
We may try to find repetitive patterns in the time series by encoding seasonal variations. There are different ways to do this.\\
One-hot encoding is a reasonable choice. Hereby, we treat seasonality as categorical variables, and use dummy variables to indicate the current time interval in the seasonal cycle. It is simple and intuitive. However, it can not really grab cyclicality, since the distance between intervals does not matter during the encoding. It means that two successive, and possibly similar, time intervals are represented by just as independent values as any two randomly chosen time intervals. Also, one-hot encoding uses an individual variable to represent each unique value, which may be inconvenient when we have a large number of time intervals. These deficiencies of the dummy variable approach lead us to another encoding method.\\
We can place the values on a single continuous scale instead of using several binary variables. By assigning increasing equidistant values to the successive time intervals, we can grab the similarity of adjacent pairs, but this encoding sees the first and the last intervals as being farthest from each other, which is a bad mistake. This may be healed by transforming the values using either the sine (\ref{eq:sine_transform}) or the cosine (\ref{eq:cosine_transform}) transformation. In order to have each interval uniquely represented, we should use both.
\begin{equation} \label{eq:sine_transform}
\tilde{x} = \sin \left ( \frac{2 \cdot \pi \cdot x}{max(x)} \right )
\end{equation}
\begin{equation} \label{eq:cosine_transform}
\tilde{x} = \cos \left ( \frac{2 \cdot \pi \cdot x}{max(x)} \right )
\end{equation}
\subsubsection{Dummy indicators}
We can use simple indicator variables for events or conditions that we consider important. Holidays are always special and are spent in unusual ways. Hence, a binary variable indicating holidays may carry information about the time series. Also, an indicator of working days or working hours could be useful.
\subsection{Feature importances}
Neural networks consist of connected simple functions of weighted sums. It is not a difficult structure, yet interpreting the meaning and role of the numerous backpropagation learnt weights is pretty hard. It is the reason for too often calling them black boxes, and not even trying to understand them. Recurrent networks are even a bit more complicated.\\
This difficulty of interpretation makes neural networks somewhat less valuable. We should naturally always prefer a simpler, more interpretable model, when having multiple choices with about the same forecasting performance. So, we should brighten the black box, at least partially.\\
A measure of variable importance could tell something about what is happening within the neural network, yet quantifying importance is not trivial. Several methods have been proposed, see \citet{gevrey2003review} or \citet{olden2004accurate} for a comparison.\\
Here we are going to use mean decrease accuracy --- a method that is usually applied to random forests (\citet{breiman2001random}). It is also called permutation accuracy. We are permuting the variables, one by one, and calculate a measure of accuracy for each deliberately corrupted model. The feature who's random permutation leads to the largest drop in accuracy is considered the most important. It is a simple, intuitive measure, and it can easily be applied not just to random forests, but to any neural network or any other supervised learning algorithm as well.\\
Two metrics will be used as measures of accuracy: $R^2$ and $MDA$. $R^2$ or coefficient of determination is a goodness of fit measure for regression. It usually ranges from 0 to 1, though it can take on negative values as well. $MDA$ or mean directional accuracy is the accuracy of the binary variable indicating the directional change of consecutive values. Mean decrease accuracy can be calculated from both accuracy measures.\\
Variables are permuted separately, so the importance scores should be interpreted and evaluated independently. Variable importances do not add up. Hence, when the same piece of information is encoded into multiple variables, we are going to calculate the maximum of the scores, and use it as the importance of that group of variables.\\
Feature importances can be estimated from both regression accuracy and directional accuracy of one-step (and maybe also multi-step) predictions on different validation sets. It means there are quite some combinations, and it is likely that not the exact same variables will prove to be important in all cases. Therefore, it is worth computing some descriptive statistics of the different calculations to get a summarized view of the actual roles of input features.\\
The importance scores are usually normalized to sum to 1.
\section{Prediction}
\subsection{Point predictions}
Our recurrent neural networks are primarily suited for one-step-ahead forecasting. We have a single output variable containing the value that immediately follows the corresponding sequence of inputs.\\
This simple framework can also be used for multi-step-ahead forecasting in an iterative (recursive) manner. When the target variable's lagged values are the only non-deterministic features, we may use the predicted values as inputs for making further predictions. It is a bit risky though, since we roll the errors forward.
\subsection{Prediction intervals}
Neural networks have an important disadvantage: they only provide point estimates. We only get some forecasted values, without any indicators of the predictions' confidence.\\
Prediction intervals contain future values with a given probability. Such interval estimates could make our forecasts more meaningful, though producing them for neural networks is not easy. We are going to use the computationally expensive method of bootstrapping.\\
The idea of bootstrapping was introduced by \citet{efron1979bootstrap}. This is actually a very simple method for estimating the quality of estimates. Bootstrapping takes resamples with replacement from the original dataset to make separate forecasts. The variability of those independent estimates is a good measure of the confidence of the estimation.\\
\citet{efron1986bootstrap} wrote a comprehensible review of bootstrapping, focusing on applications. They state that the bootstrap method has very nice properties for estimating standard errors, and computational intensity, being its main weakness, is less and less of a case as computation is getting cheaper and faster. Efron and Tibshirani also note that the bootstrap can be used to construct confidence intervals in an automatic way, while alternative methods require several tricks to perform satisfactorily.\\
Bootstrapping is not only useful for quantifying the confidence in our predictions, but also for improving the forecasts themselves. Ensemble methods combine multiple algorithms to deliver better predictions. Bootstrapping is one way to construct an ensemble.\\
There are 2 main approaches to ensemble learning (\citet{dietterich2002ensemble}). Either we can create models in a coordinated fashion, and take a weighted vote of the components, or we can construct independent models with a diverse set of results, then combine them to let the disagreements cancel out. Bootstrap resampling can be used to construct an ensemble of this latter type.\\
Bagging or bootstrap aggregating is an ensemble method that trains a learning algorithm on multiple independent bootstrap samples, and aggregates the resulting predictors by voting or averaging. It was introduced by \citet{breiman1996bagging}. This simple ensemble can improve unstable single predictors, but it can slightly degrade the performance of more stable ones. Hence, applying bagging is not always a good idea, but it can work very well for some learning algorithms. \citet{dietterich2002ensemble} remarks that since the generalization ability of neural networks is very good, they may benefit less from ensemble methods. Still, we may hope that bagging can bring some improvement to our point predictions, but the main reason for applying the method is the construction of prediction intervals.\\
Bootstrapping has been applied to compute confidence and prediction intervals for neural networks by, e.g., \citet{paass1993assessing}, \citet{heskes1997practical}, \citet{carney1999confidence}. \citet{khosravi2011comprehensive} compared four leading techniques for constructing prediction intervals, including the bootstrap.  They conclude that there is no method that outperforms all the others in all respects. \citet{tibshirani1996comparison} compared two different bootstraps to two other methods for estimating the standard error of neural network predictions. \citet{tibshirani1996comparison} concludes that the bootstraps perform best.\\
Here we are going to use a method similar to the one proposed by \citet{heskes1997practical}.\\
We take bootstrap samples of sequences, rather than individual observations. A separate RNN is trained on each bootstrap sample. Our final (point) prediction is a simple average of all individual model predictions.\\
We must make a distinction between confidence intervals and prediction intervals. These two are easily confused.\\
Confidence intervals quantify how well we can approximate the true regression. The confidence is in the estimate of the regression --- the mean of the target distribution.\\
To compute the intervals, we first construct the bagged estimator by averaging the resulting estimates of all bootstrap runs (\ref{eq:pi_bagging}).\\
\begin{equation} \label{eq:pi_bagging}
\hat{y}_i = \frac{1}{n_{run}} \sum_{run=1}^{n_{run}} \hat{y}^{run}_i
\end{equation}
Now we have the center, and we just need to find the variance (\ref{eq:pi_variance_outcomes}) and an appropriate value from the t-distribution to calculate the endpoints of the confidence interval (\ref{eq:pi_confidence_interval}). We assume that the true regression follows a normal distribution given the estimate.
\begin{equation} \label{eq:pi_variance_outcomes}
\sigma_{\hat{y}_i}^2 = \frac{1}{n_{run}-1} \sum_{run=1}^{n_{run}} (\hat{y}_i^{run}-\hat{y}_i)^2
\end{equation}
\begin{equation} \label{eq:pi_confidence_interval}
%\hat{y}_i - t_{conf} \cdot \sigma_{\hat{y}_i} \leq y_i \leq \hat{y}_i + t_{conf} \cdot \sigma_{\hat{y}_i}%?
%\hat{y}_i \pm  t_{conf} \cdot \sigma_{\hat{y}_i}
CI_i = [\hat{y}_i - t_{conf} \cdot \sigma_{\hat{y}_i}, \: \hat{y}_i + t_{conf} \cdot \sigma_{\hat{y}_i}]
\end{equation}
Prediction intervals quantify how well we can approximate the target values. This measure is more important in practice, but it is a bit more difficult to estimate. While the construction of confidence intervals required nothing more than the means and standard deviations of our resampled estimates, here we need some more sophisticated computations to estimate the noise variance of the regression (\ref{eq:pi_variance_errors}).\\
\begin{equation} \label{eq:pi_variance_errors}
\sigma_{\hat{\epsilon}}^2 \simeq E[(y-\hat{y})^2] - \sigma_{\hat{y}}^2%\simeq 
\end{equation}
We train another network with almost the same structure as the one used for making the time series predictions, and use it to predict the remaining residuals (\ref{eq:pi_residuals}) from the input values.\\
\begin{equation} \label{eq:pi_residuals}
r_i^2 = max((y_i-\hat{y}_i)^2-\sigma_{\hat{y}_i}^2,0)
\end{equation}
This residual predictor is trained on the validation set of observations randomly left out in the current bootstrap run. It is a smart data recycling method. \citet{heskes1997practical} proposed loglikelihood as the loss function, we apply the similar formula (\ref{eq:pi_loss}) used by \citet{khosravi2011comprehensive}. The output activation of this neural network is exponential, so that all predicted error variances are positive.\\
\begin{equation} \label{eq:pi_loss}
L = \frac{1}{2} \sum_{i=1}^{n}(ln(\sigma_{\hat{\epsilon}_i}^2)+\frac{r_i^2}{\sigma_{\hat{\epsilon}_i}^2})
\end{equation}
The output variances and the neural network's predictions are added together (\ref{eq:pi_variance}) to yield estimates of the variance of real interest --- the distance of target values from our bagged forecast estimates (\ref{eq:pi_prediction_interval}).
\begin{equation} \label{eq:pi_variance}
\sigma_i^2 = \sigma_{\hat{y}_i}^2 + \sigma_{\hat{\epsilon}_i}^2
\end{equation}
\begin{equation} \label{eq:pi_prediction_interval}
PI_i = [\hat{y}_i - t_{conf} \cdot \sigma_i, \: \hat{y}_i + t_{conf} \cdot \sigma_i]
\end{equation}
\section{Validation}
\subsection{Point forecasts}
We are going to obtain a train and a test set by splitting the time series at one point. It means that we always test the future. We would expect the algorithm to tell the future, so this choice of validation is natural for such forecasting problems. Yet, the test set consists of a single time period, so this method may not be entirely sufficient for evaluating the model performance.\\
Bootstrapping provides an alternative validation set. An average bootstrap sample contains about 63.2\% of the individual observations, or in our case, it contains about 63.2\% of all available data subsequences. The remaining subsequences do not participate in the training process, so we may use them for validation purposes.\\
We are going to call the bootstrap left-out dataset validation set, and the future dataset test set --- just for the sake of distinction. They have the same evaluation purpose.\\
Both sets will be used to evaluate the one-step-ahead forecasting ability of our recurrent neural networks. The separate test set, being a complete chronologically ordered series of subseries, may also be used for iterative multi-step-ahead forecasting.\\
Regression and classification metrics are going to be applied in order to evaluate the forecasted values and the predicted changes of direction as well.\\
We are calculating root mean squared error (\ref{eq:metrics_rmse}), symmetric mean absolute percentage error (\ref{eq:metrics_smape}), coefficient of determination (\ref{eq:metrics_r2}), mean absolute error (\ref{eq:metrics_mae}) and median absolute error (\ref{eq:metrics_medae}) regression metrics to measure the forecast fit. We are also calculating the accuracy ($MDA$) (\ref{eq:metrics_accuracy}), precision (\ref{eq:metrics_precision}), recall (\ref{eq:metrics_recall}) and F1 (\ref{eq:metrics_f1}) classification scores to evaluate the directional forecasts. The classification metrics are computed on the following variables: $\tilde{y}_t = 1_{y_t - y_{t-1} > 0}$ and $\tilde{\hat{y}}_t = 1_{\hat{y}_t - y_{t-1} > 0}$.\\
\begin{equation} \label{eq:metrics_rmse}
RMSE(y, \hat{y}) = \sqrt{\frac{1}{n}\sum_{i=1}^{n}(y_i - \hat{y}_i)^2}
\end{equation}
%\begin{equation} \label{eq:metrics_mape}
%MAPE(y, \hat{y}) = \frac{100}{n} \sum_{i=1}^{n}|\frac{y_i-\hat{y}_i}{y_i}|
%\end{equation}
\begin{equation} \label{eq:metrics_smape}
SMAPE(y, \hat{y}) = \frac{100}{n} \sum_{i=1}^{n}\frac{|y_i - \hat{y}_i|}{([y_i] + |\hat{y}_i|)/2}
\end{equation}
\begin{equation} \label{eq:metrics_r2}
R^2(y, \hat{y}) = 1 - \frac{\sum_{i=1}^{n}(y_i - \hat{y}_i)^2}{\sum_{i=1}^{n}(y_i - \bar{y})^2}
\end{equation}
\begin{equation} \label{eq:metrics_mae}
MAE(y, \hat{y}) = \frac{1}{n} \sum_{i=1}^{n}|y_i - \hat{y}_i|
\end{equation}
\begin{equation} \label{eq:metrics_medae}
MedAE(y, \hat{y}) = median(|y_1 - \hat{y}_1|, \dots , |y_n - \hat{y}_n|)
\end{equation}
\begin{equation} \label{eq:metrics_accuracy}
accuracy(\tilde{y}, \tilde{\hat{y}}) = \frac{1}{n} \sum_{i=1}^{n}1_{\tilde{y}_i=\tilde{\hat{y}}_i}
\end{equation}
\begin{equation} \label{eq:metrics_precision}
precision(\tilde{y}, \tilde{\hat{y}}) = \frac{ \sum_{i=1}^{n}1_{\tilde{y}_i=1 \:  and \: \tilde{\hat{y}}_i=1}}{ \sum_{i=1}^{n}1_{\tilde{\hat{y}}_i=1}}
\end{equation}
\begin{equation} \label{eq:metrics_recall}
recall(\tilde{y}, \tilde{\hat{y}}) = \frac{\sum_{i=1}^{n}1_{\tilde{y}_i=1 \: and \: \tilde{\hat{y}}_i=1}}{\sum_{i=1}^{n}1_{\tilde{y}_i=1}}
\end{equation}
\begin{equation} \label{eq:metrics_f1}
F1(\tilde{y}, \tilde{\hat{y}}) = 2 \cdot \frac{precision \cdot recall}{precision + recall}
\end{equation}
$MSE$ is used as the loss function during the training process. Applying other evaluation metrics is also reasonable, since they all have different properties and interpretation. $MAE$ (the average of absolute errors) is easier to interpret than the square root of $MSE$, $MedAE$ is more robust to outliers, $R^2$ score measures the proportion of variance explained by the model, $SMAPE$ measures the error in percentage terms, ranging from 0 to 200\%.\\
Classification $accuracy$ score simply measures the proportion of changes whose direction we guessed right. $Precision$, $recall$ and $F1$ are binary metrics. An increase in the target variable is now treated as the positive class. Hence, in our case, $precision$ is the proportion of forecasted rises that were right, and $recall$ is the proportion of actual rises that we guessed right. $F1$ score is the harmonic mean of precision and recall. We may get a better view of the forecasts' quality by  using these various metrics together. Also, we can draw a confusion matrix to evaluate all kinds of errors and correct forecasts at once.
\subsection{Interval forecasts}
The quality of prediction intervals should also be evaluated. We use the same prediction interval assessment measures as \citet{khosravi2011comprehensive}.\\
$PICP$ or prediction interval coverage probability (\ref{eq:pi_picp}) is the proportion of observations that fall into the interval. The wider the interval, the higher the coverage --- it is reasonable to quantify the size of the interval as well. $MPIW$ or mean prediction interval width (\ref{eq:pi_mpiw}) does exactly that. $NMPIW$ or normalized mean prediction interval width (\ref{eq:pi_nmpiw}) is the width normalized by the range of the target variable. This metric allows for comparisons across datasets. $CWC$ or coverage width-based criterion (\ref{eq:pi_cwc}) is a combined metric that takes into account both coverage and width. It has 2 hyperparameters that we can set: $\mu$ corresponds to the nominal confidence level, while $\eta$ magnifies the difference between $\mu$ and $PICP$. The lower $CWC$ the higher quality.
\begin{equation} \label{eq:pi_picp}
PICP = \frac{1}{n} \sum_{i=1}^{n}1_{y_i \in [L_i, H_i]}
\end{equation}
\begin{equation} \label{eq:pi_mpiw}
MPIW = \frac{1}{n}\sum_{i=1}^{n}(U_i-L_i)
\end{equation}
\begin{equation} \label{eq:pi_nmpiw}
NMPIW = \frac{MPIW}{R}
\end{equation}
\begin{equation} \label{eq:pi_cwc}
CWC = NMPIW (1 + 1_{PICP<\mu} exp(-\eta(PICP-\mu)))
\end{equation}
\section{Empirical study}
Our forecasting framework was implemented in Python (\citet{van1995python}). This section presents an example application on the Bike Sharing Dataset (\citet{fanaee2014event}) available in the UCI Machine Learning Repository (\citet{Dua:2017}).
\subsection{Features}
\begin{figure}
\centering
%same folder
\includegraphics[width=\textwidth,height=\textheight,keepaspectratio]{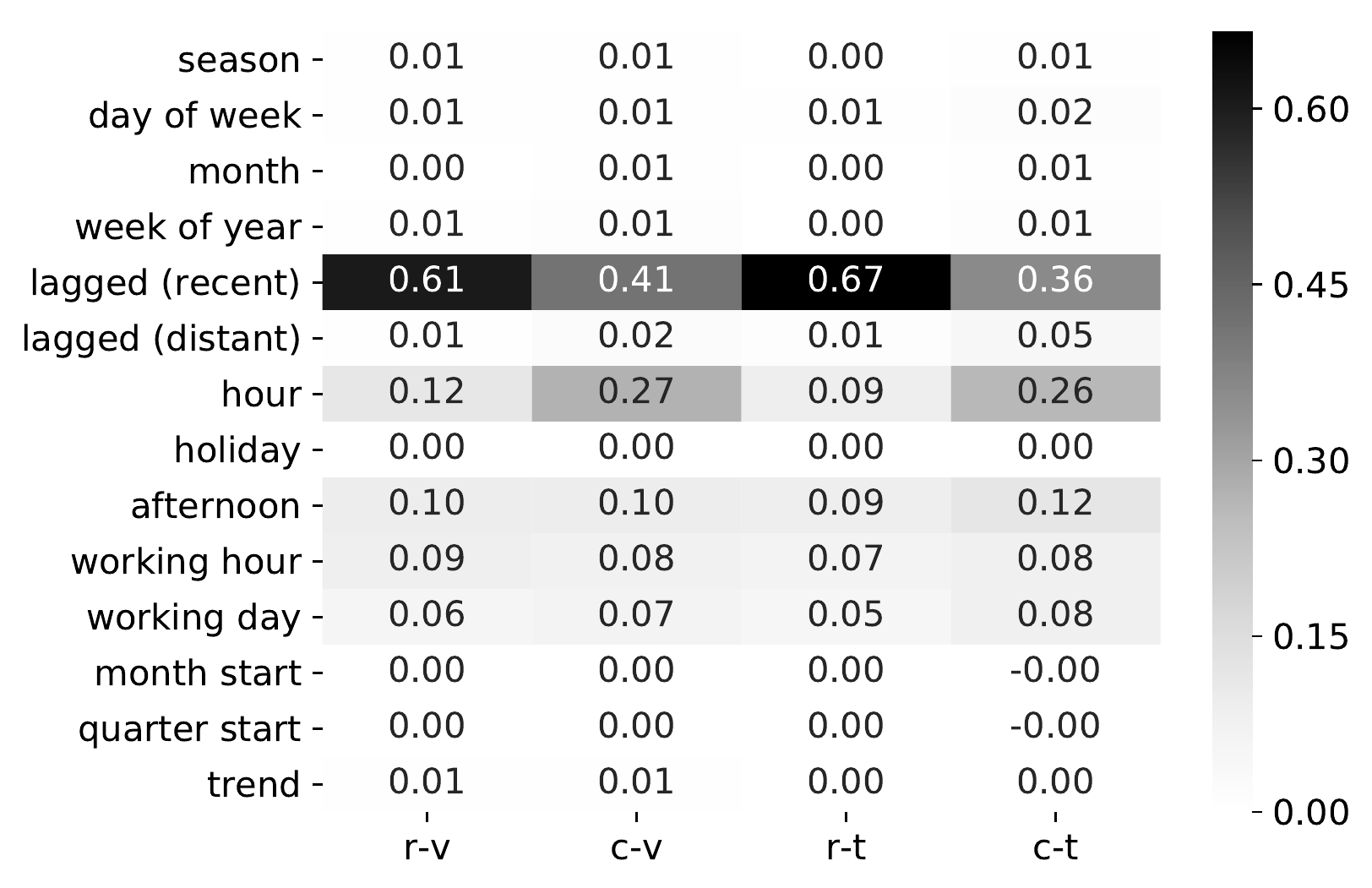}
\caption{Feature importances, detailed [LSTM]}
\label{fig:importance_all_lstm}
\end{figure}

\begin{figure}
\centering
\includegraphics[width=\textwidth,height=\textheight,keepaspectratio]{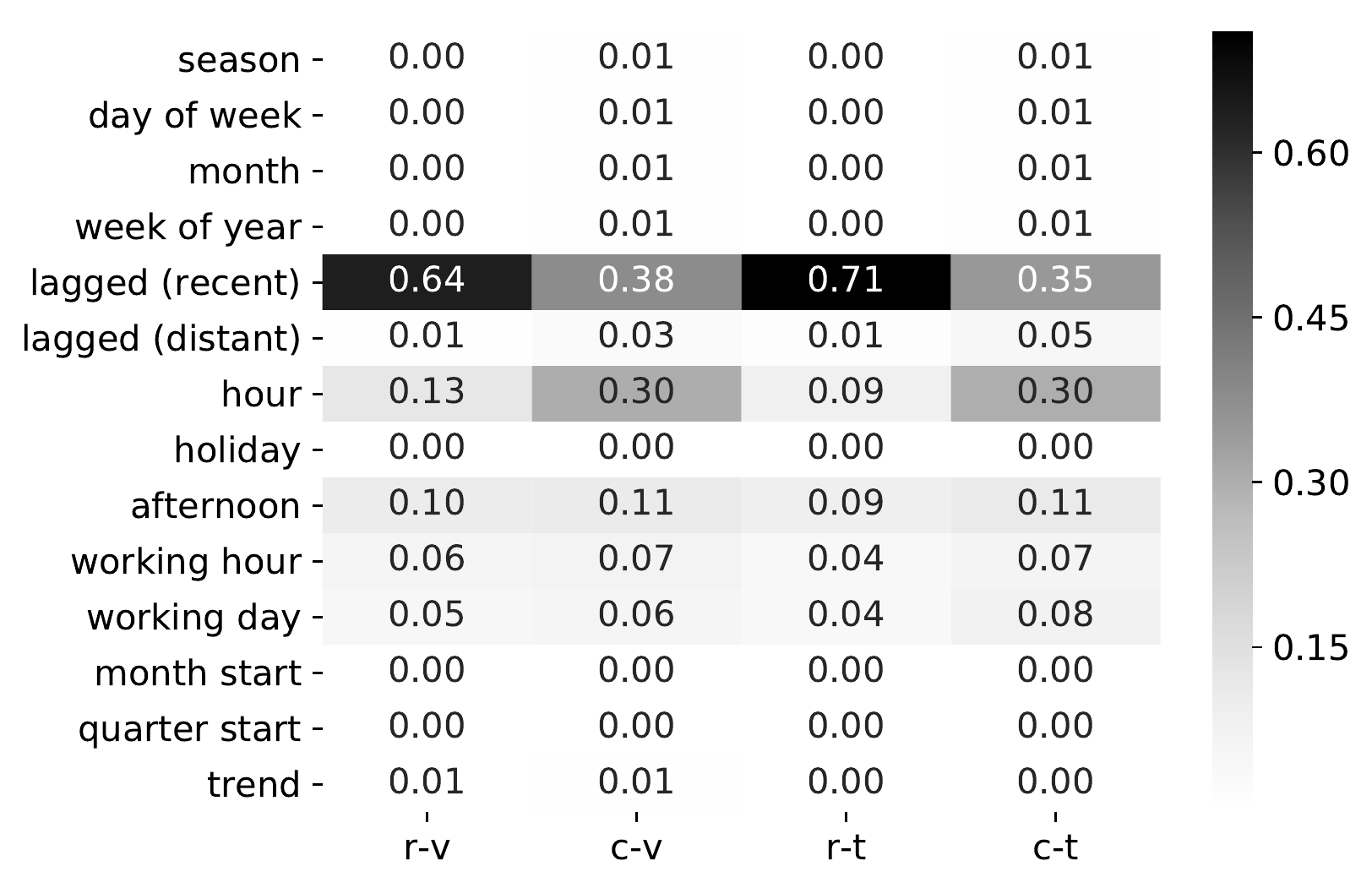}
\caption{Feature importances, detailed [GRU]}
\label{fig:importance_all_gru}
\end{figure}

\begin{figure}
\centering
\includegraphics[width=\textwidth,height=\textheight,keepaspectratio]{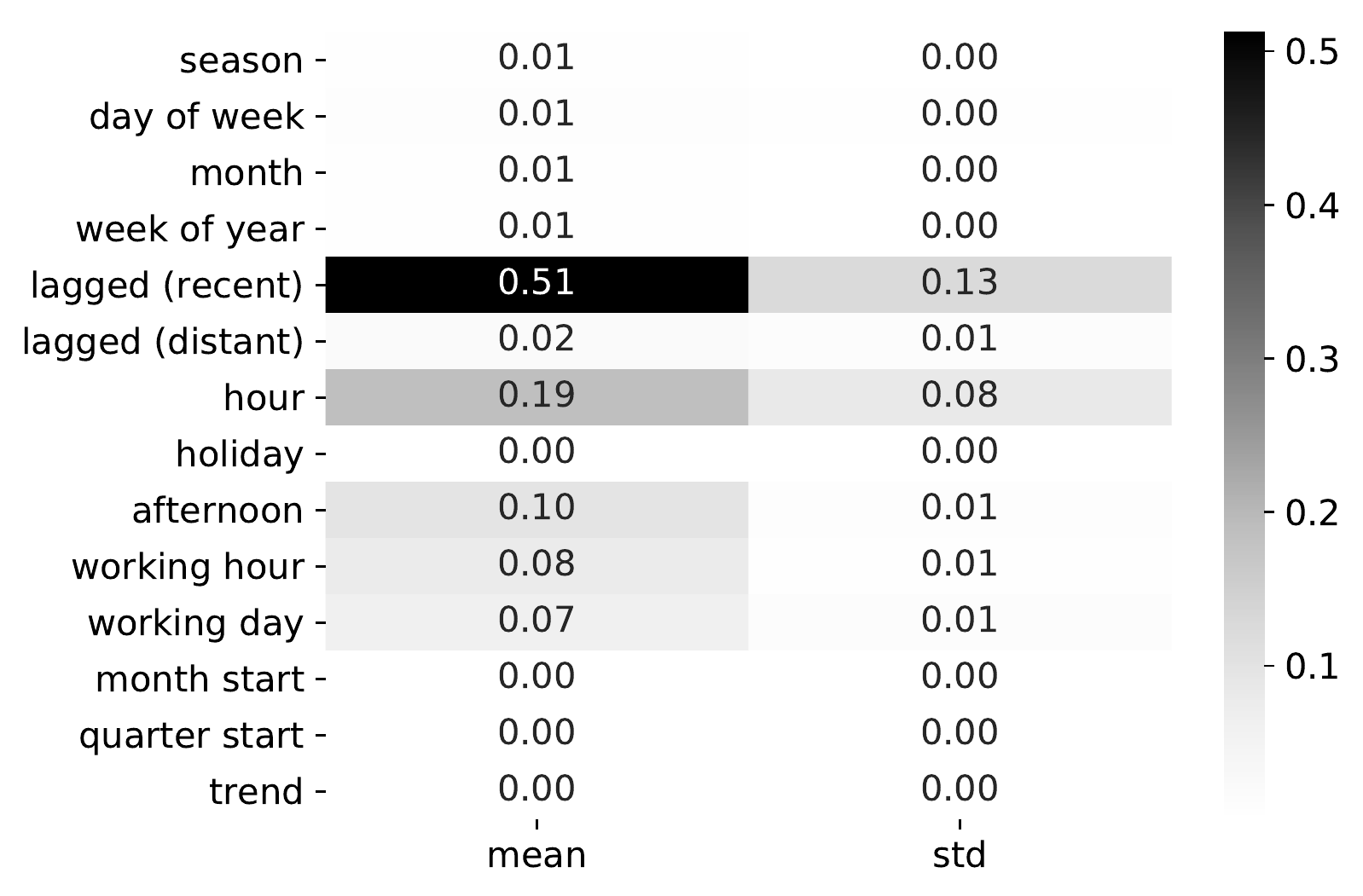}
\caption{Feature importances, summary [LSTM]}
\label{fig:importance_desc_lstm}
\end{figure}

\begin{figure}
\centering
\includegraphics[width=\textwidth,height=\textheight,keepaspectratio]{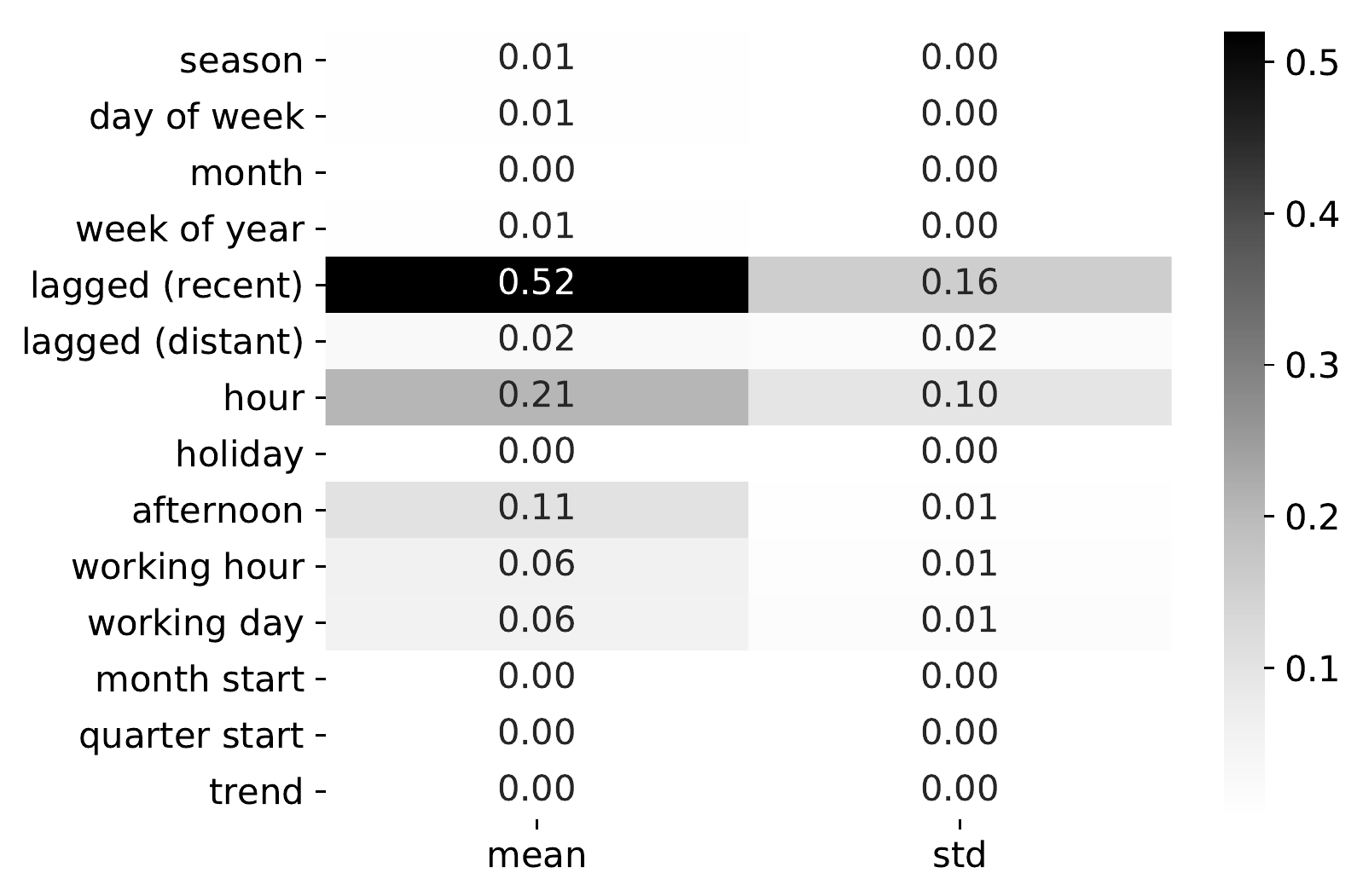}
\caption{Feature importances, summary [GRU]}
\label{fig:importance_desc_gru}
\end{figure}

The dataset is available in an hourly resolution, which allows us to construct several seasonal variables. It contains counts of bike rentals for Capital Bikeshare at Washington, D.C., USA for 2 years. We disregard the available weather information, and use time-determined features only. The pandas library (\citet{mckinney2010data}) was used during the data preparation.\\
Cyclical features were encoded using the sine and cosine transformations. The following time components are encoded: season of year, month of year, week of year, day of week and hour of day. Using all these features together is probably a bit redundant, but hopefully the neural network is smart enough to handle it. Some binary variables are used to indicate if the time is afternoon, working hour, holiday, working day, start of quarter or start of month. Lagged values of the target variable are also used as inputs: the preceding 3 timesteps' values, and lagged values of 24, 48 and 168 hours.\\
Each feature was scaled between zero and one using minmax scaler.\\
Variable importances were calculated on the bootstrap left-out validation set and on the separate test set as well, using the single step forecasts. Two accuracy metrics were applied: $R^2$ as a measure of goodness-of-fit, and $MDA$ as a measure of directional accuracy. This setting led to 4 different estimates of variable importance (Figures \ref{fig:importance_all_lstm} and \ref{fig:importance_all_gru}). Those estimates were then averaged into a single list of importances (Figures \ref{fig:importance_desc_lstm} and \ref{fig:importance_desc_gru}).\\
Feature importances are displayed with the following notation: r -- regression metric ($R^2$); c -- classification metric ($accuracy$); v -- bootstrap left-out validation set; t -- future test set. These importance scores were only computed for one-step forecasts, since shuffling the values of the iterative multi-step forecasts was inconvenient.\\
%maybe
Lagged values were divided into 2 groups: recent values (lags of 1, 2 and 3 hours) and distant values (lags of 24, 48 and 168 hours). Naturally, the sine and cosine transformed values of cyclical features were also treated together. The maximum mean decrease accuracies are reported for these grouped variables.\\
Recent lagged values constitute the most important group of variables. They are especially important for the value forecasts (regression). Seasonal variables also seem to play an important role --- intraday features like the hour of day, afternoon or working hour got large importance scores. The trend variable does not seem to have that much importance, though it was evident during model building attempts that it can add to the RNNs' forecasting capability.\\
Variable importances were pretty similar for the LSTM and GRU networks. We can hardly see any disagreements in the feature rankings.
\subsection{Predictions}
\begin{figure}
\centering
\begin{subfigure}{.5\textwidth}
  \centering
  \includegraphics[width=1.\linewidth]{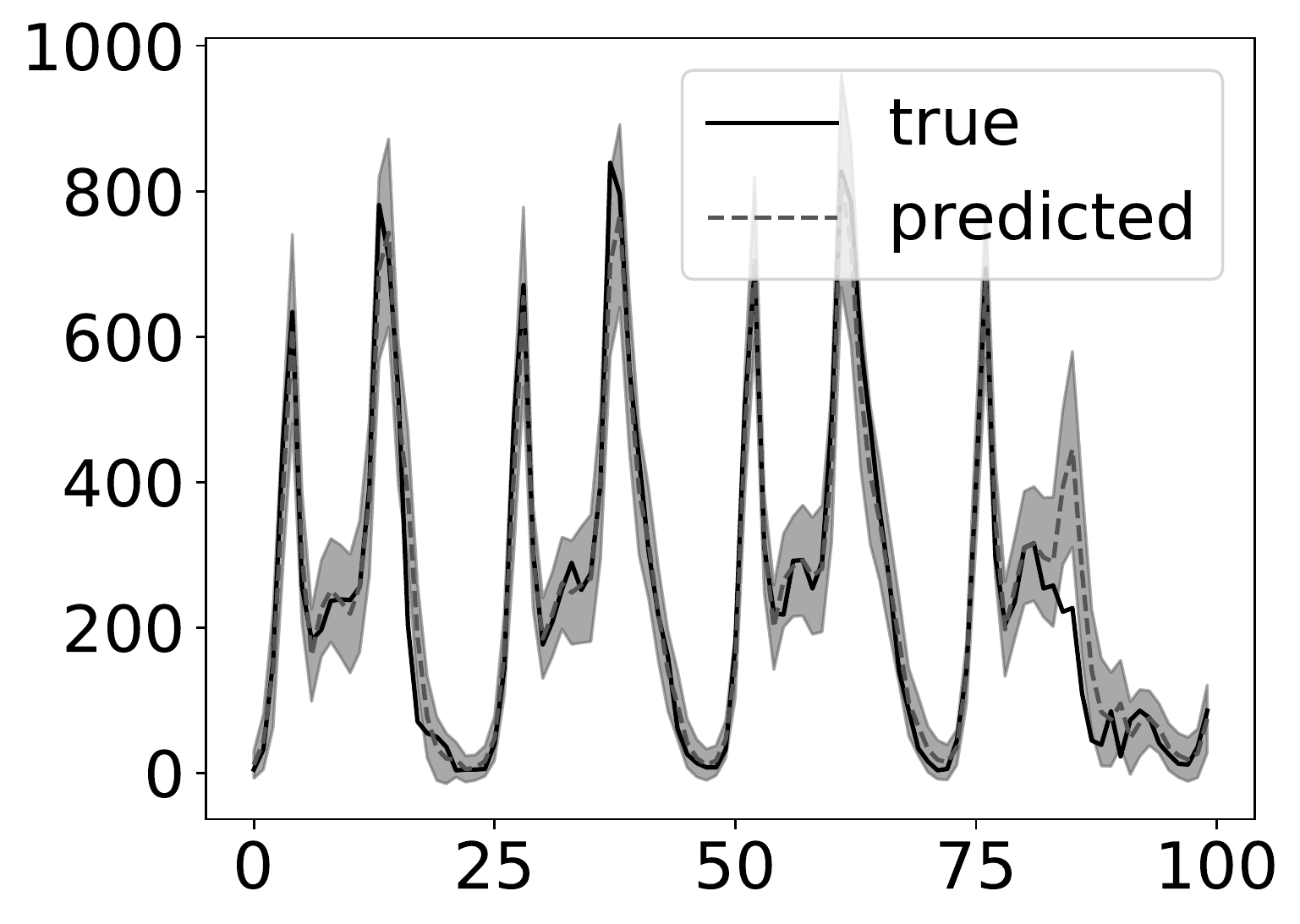}
  \caption{One-step forecasts}
  \label{fig:preds_lstm_one}
\end{subfigure}%
\begin{subfigure}{.5\textwidth}
  \centering
  \includegraphics[width=1.\linewidth]{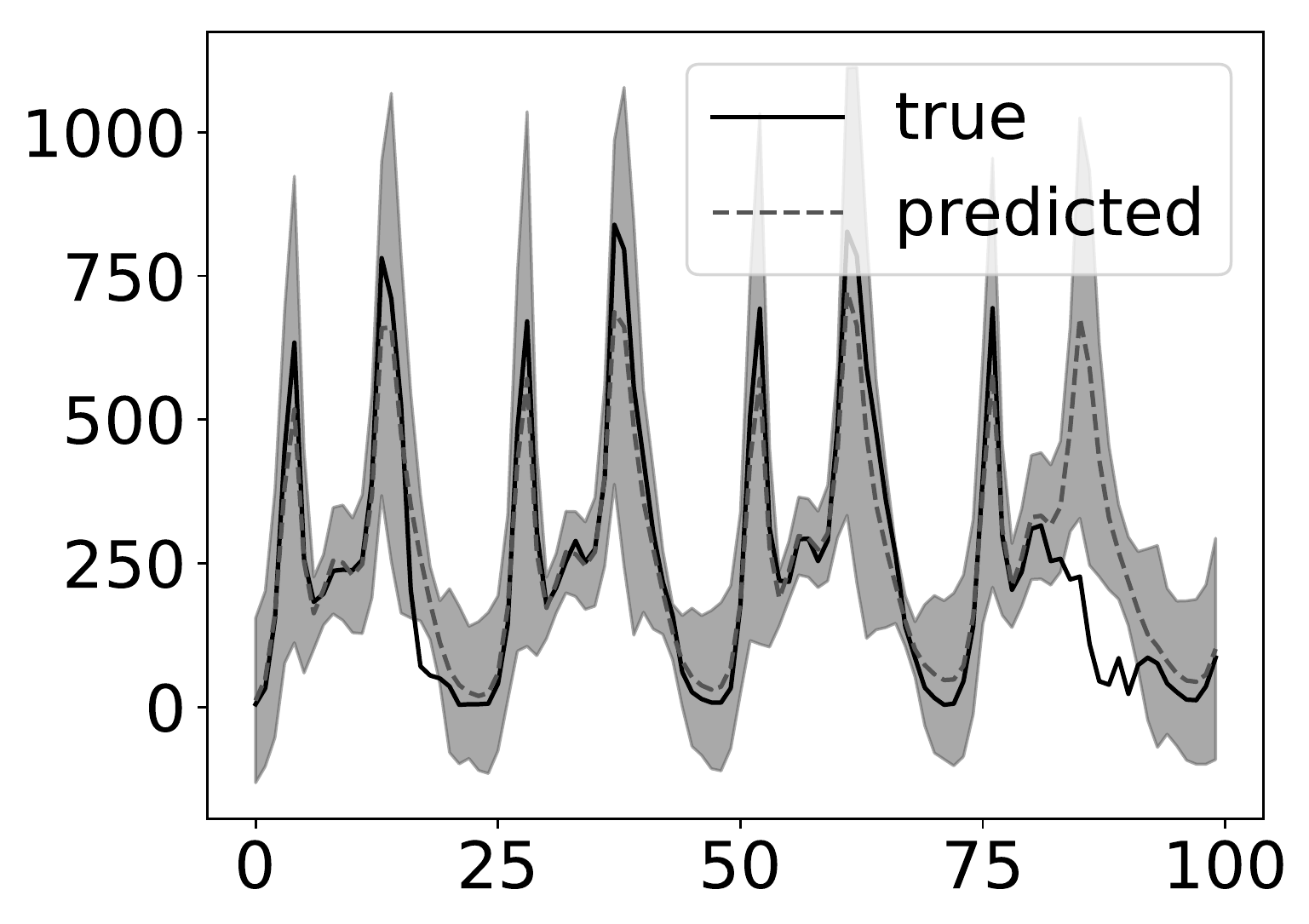}
  \caption{Multi-step forecasts}
  \label{fig:preds_lstm_multi}
\end{subfigure}
\caption{First 100 steps of test set forecasts with 90\% prediction intervals [LSTM]}
\label{fig:preds_lstm}
\end{figure}

\begin{figure}
\centering
\begin{subfigure}{.5\textwidth}
  \centering
  \includegraphics[width=1.\linewidth]{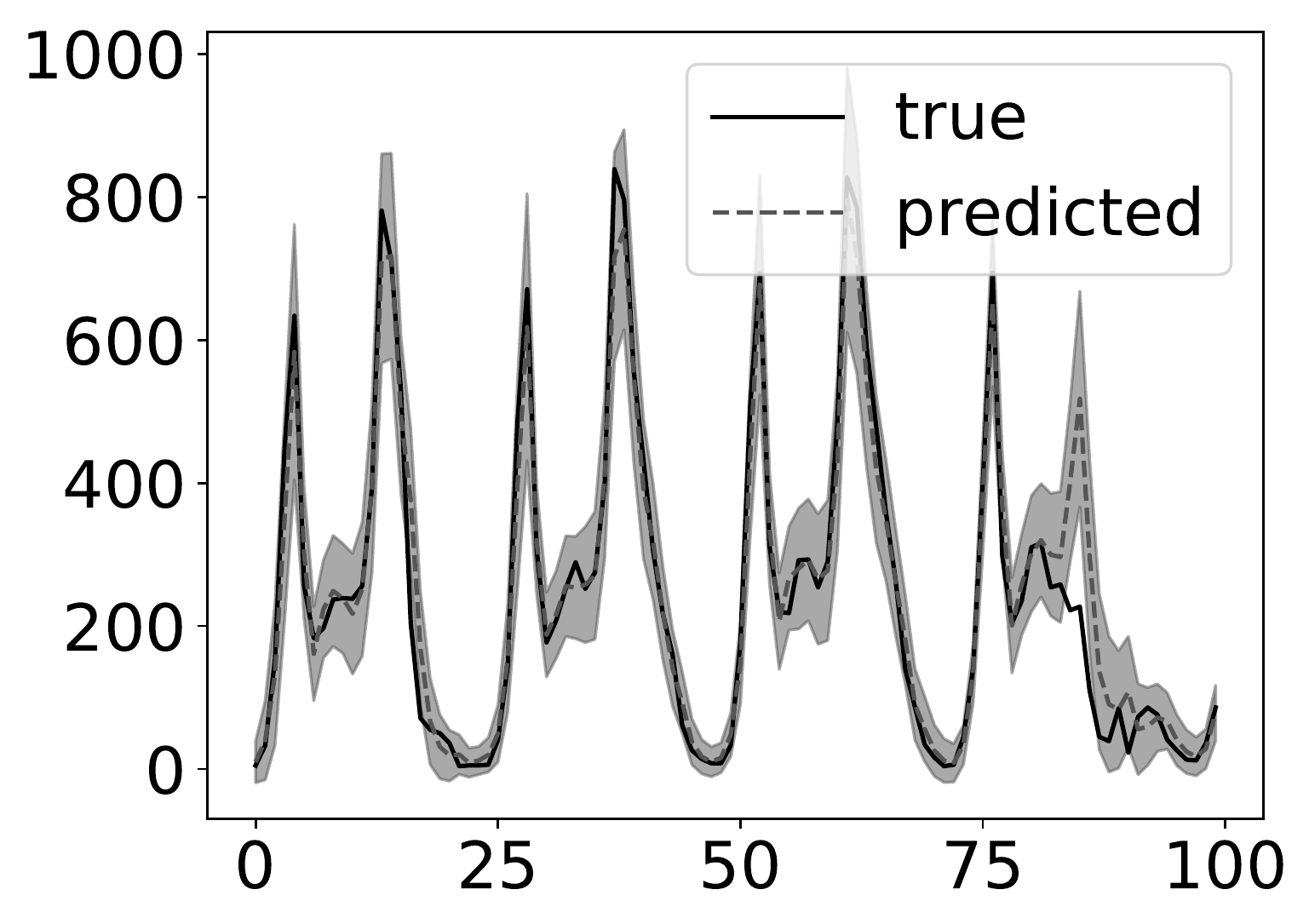}
  \caption{One-step forecasts}
  \label{fig:preds_gru_one}
\end{subfigure}%
\begin{subfigure}{.5\textwidth}
  \centering
  \includegraphics[width=1.\linewidth]{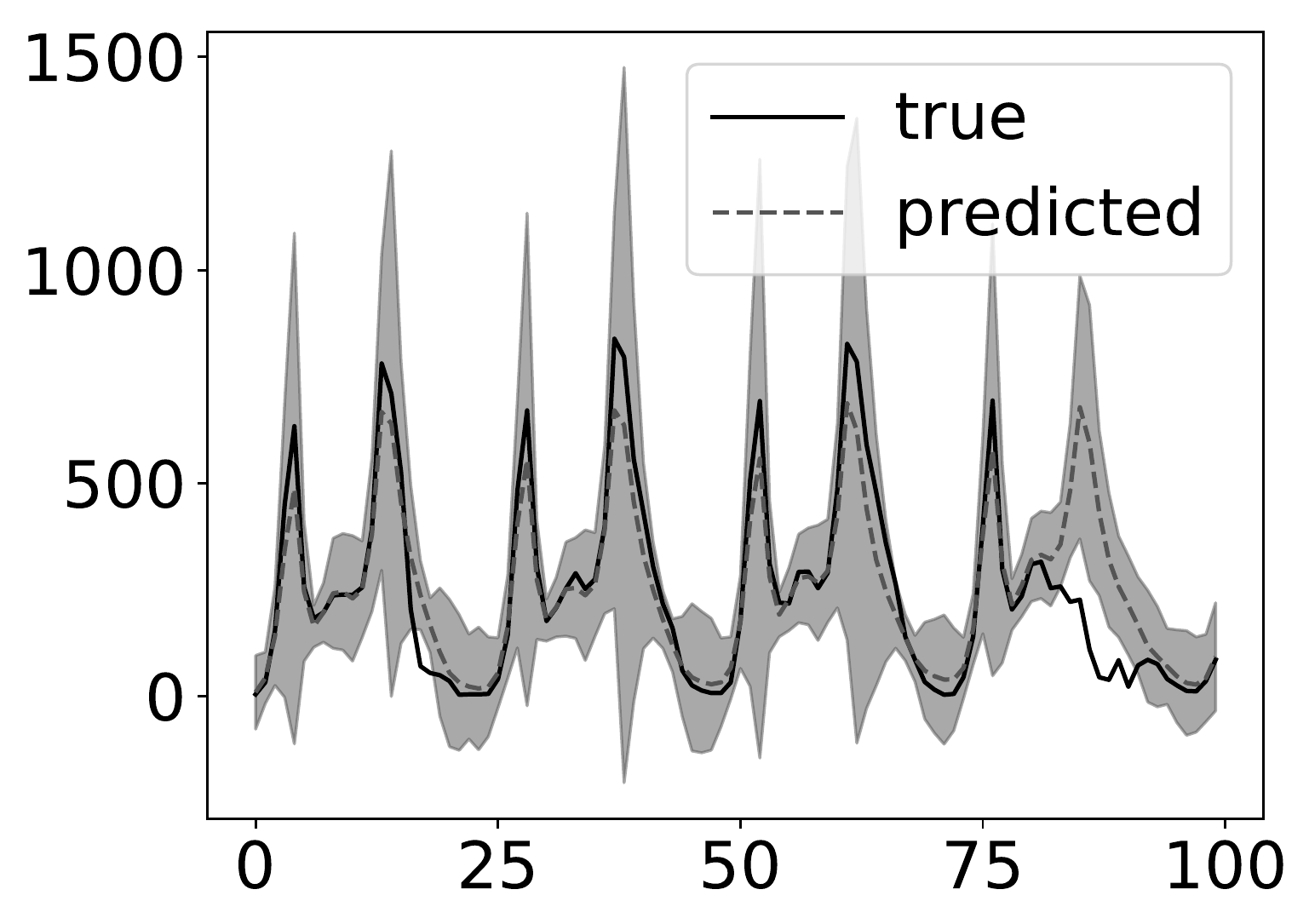}
  \caption{Multi-step forecasts}
  \label{fig:preds_gru_multi}
\end{subfigure}
\caption{First 100 steps of test set forecasts with 90\% prediction intervals [GRU]}
\label{fig:preds_gru}
\end{figure}

Our recurrent neural networks consist of a one-layer LSTM/GRU of 32 units, followed by a dense layer of a single unit with a linear activation. A dropout of 0.5 was applied to the non-recurrent connections. The learning rate was set to 0.001. The batch size and the number of epochs were 128. The mean squared error loss function was minimized using the Adam optimizer (\citet{kingma2014adam}). 16-step unrolled sequences were fed to the algorithm.\\
These hyperparameters were not optimized. It is just a reasonable setting for showcasing the method on the given dataset.\\
The model was trained on 70\% of the available data, the last 30\% was used as a test set. 50 bootstrap samples were taken from the training set.\\
Forecasts were made with approximate 90\% prediction intervals. Multi-step forecasts were generated for the test set only. Some predictions are shown in Figure \ref{fig:preds_lstm} and Figure \ref{fig:preds_gru}.\\
The recurrent neural networks were built and trained using Keras (\citet{chollet2015keras}) with TensorFlow backend (\citet{tensorflow2015-whitepaper}).
\subsection{Evaluation}
\begin{figure}
\centering
\begin{subfigure}{.5\textwidth}
  \centering
  \includegraphics[width=.8\linewidth]{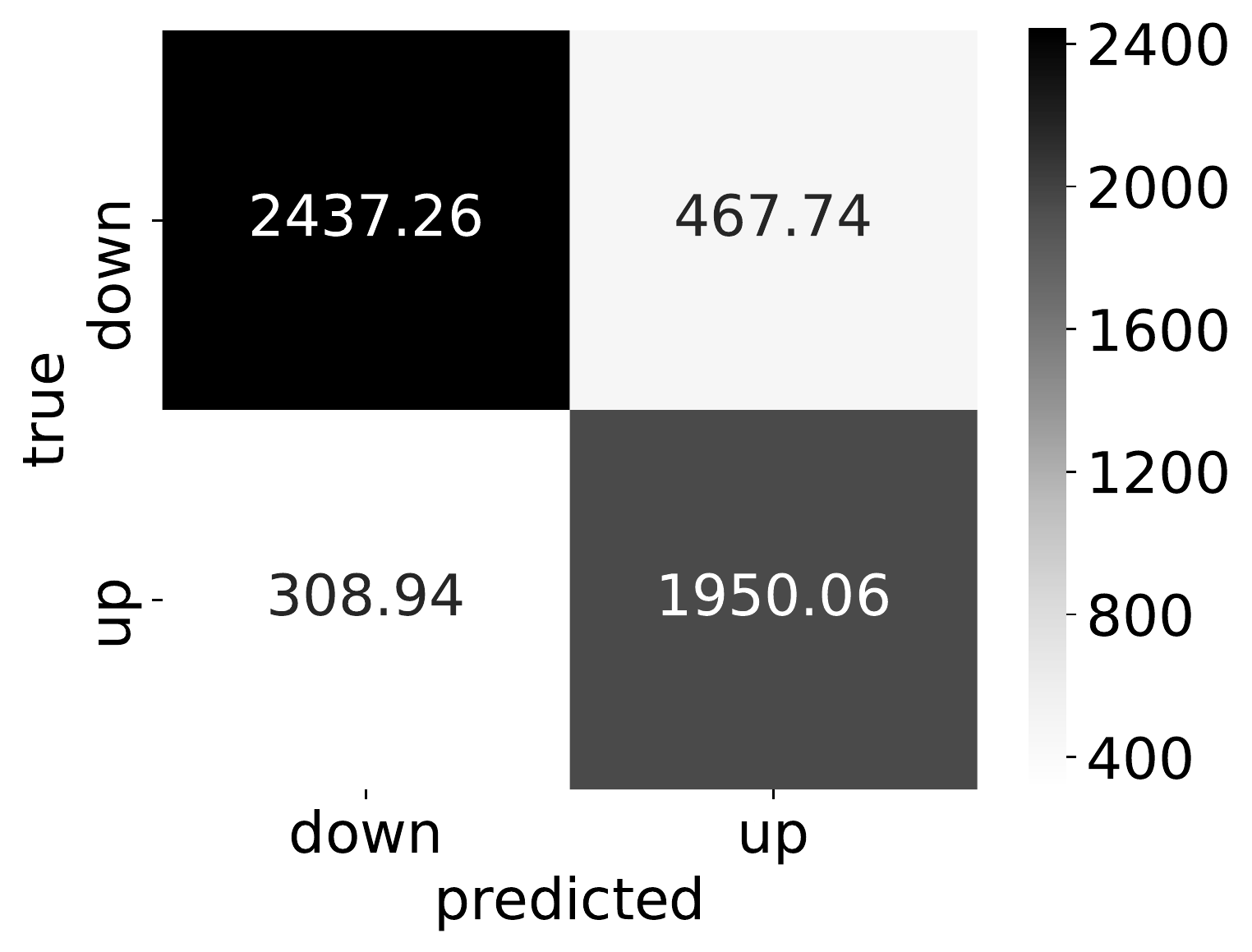}
  \caption{One-step, LSTM}
  \label{fig:conf_mtx_avg_lstm_one}
\end{subfigure}%
\begin{subfigure}{.5\textwidth}
  \centering
  \includegraphics[width=.8\linewidth]{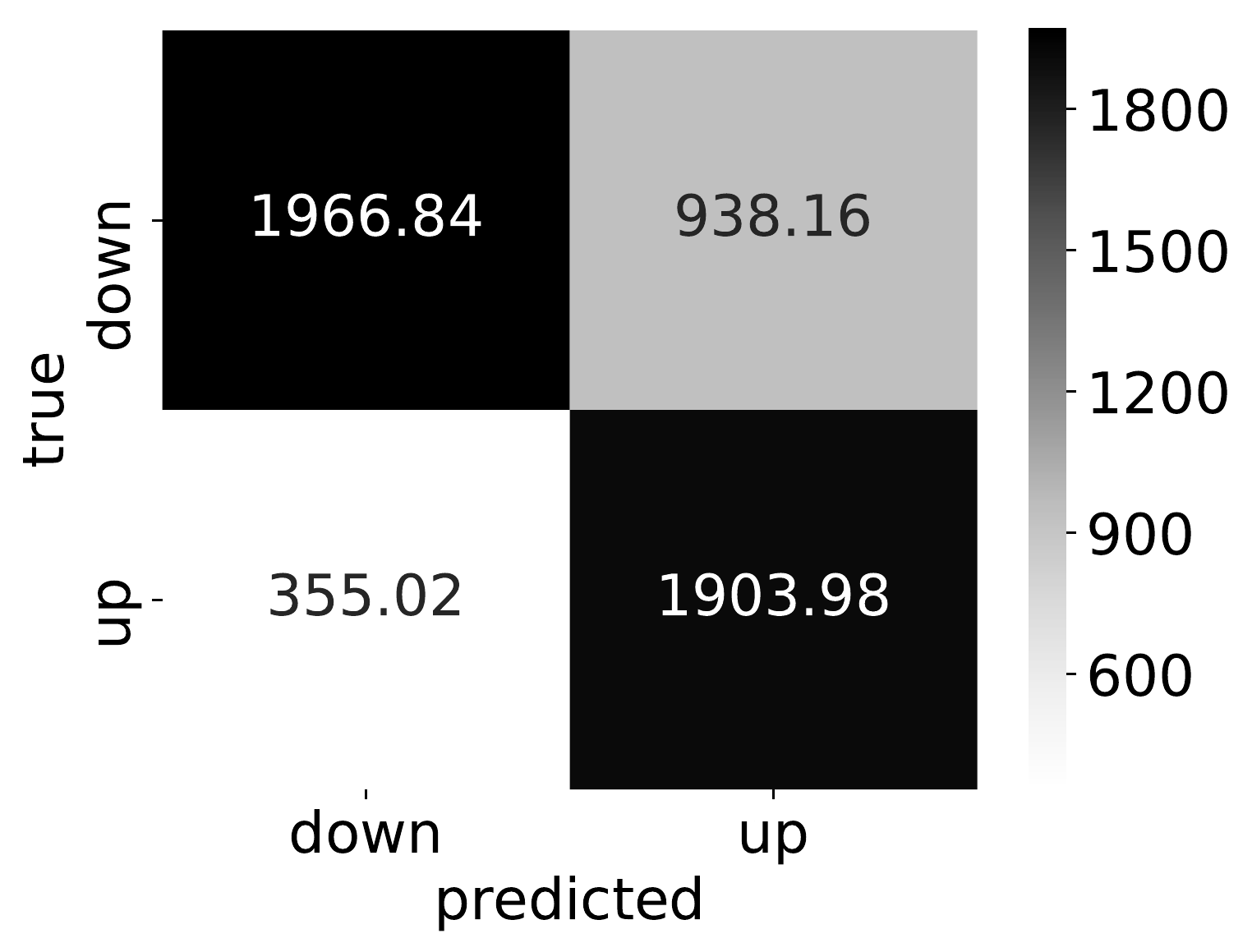}
  \caption{Multi-step, LSTM}
  \label{fig:conf_mtx_avg_lstm_multi}
\end{subfigure}
\begin{subfigure}{.5\textwidth}
  \centering
  \includegraphics[width=.8\linewidth]{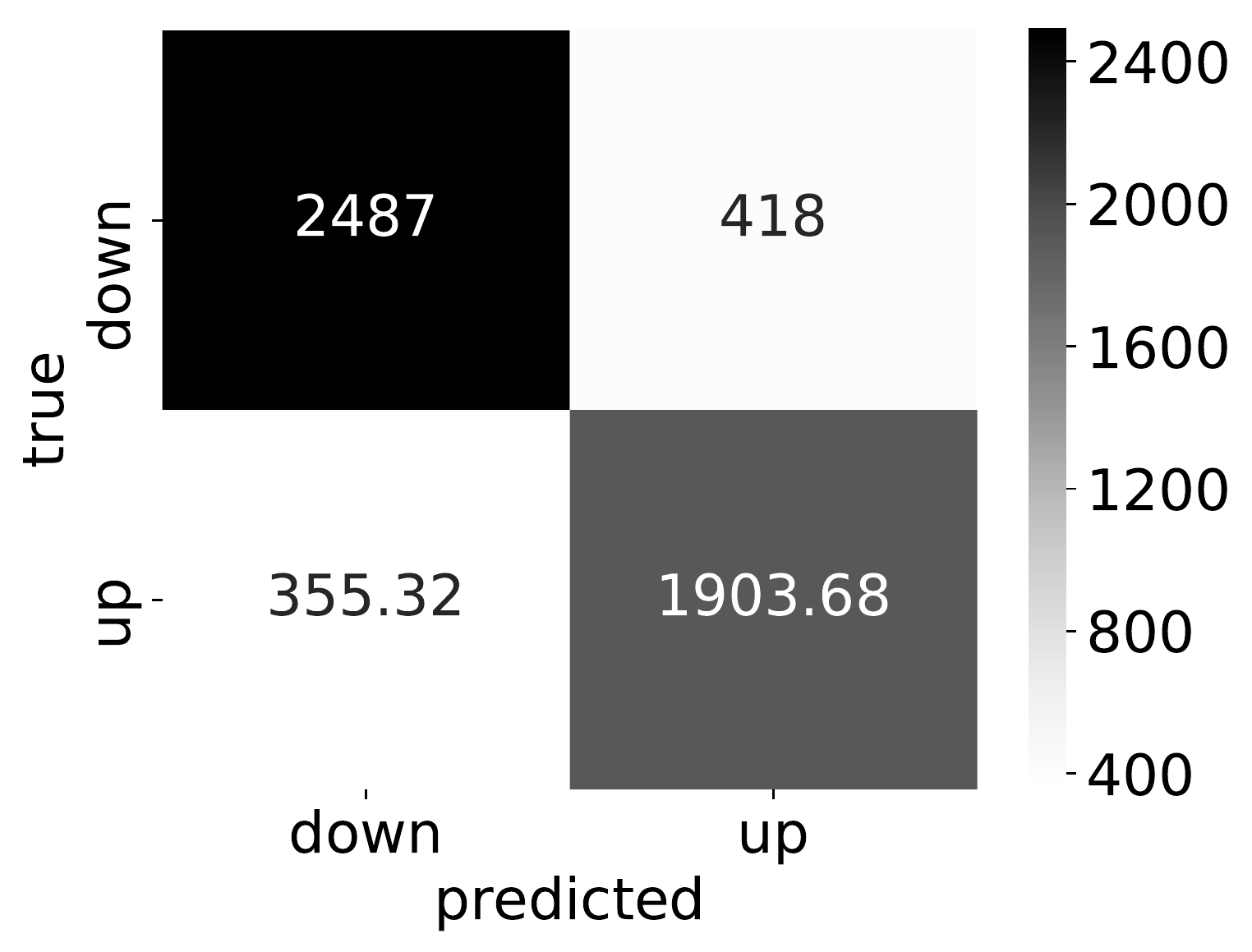}
  \caption{One-step, GRU}
  \label{fig:conf_mtx_avg_gru_one}
\end{subfigure}%
\begin{subfigure}{.5\textwidth}
  \centering
  \includegraphics[width=.8\linewidth]{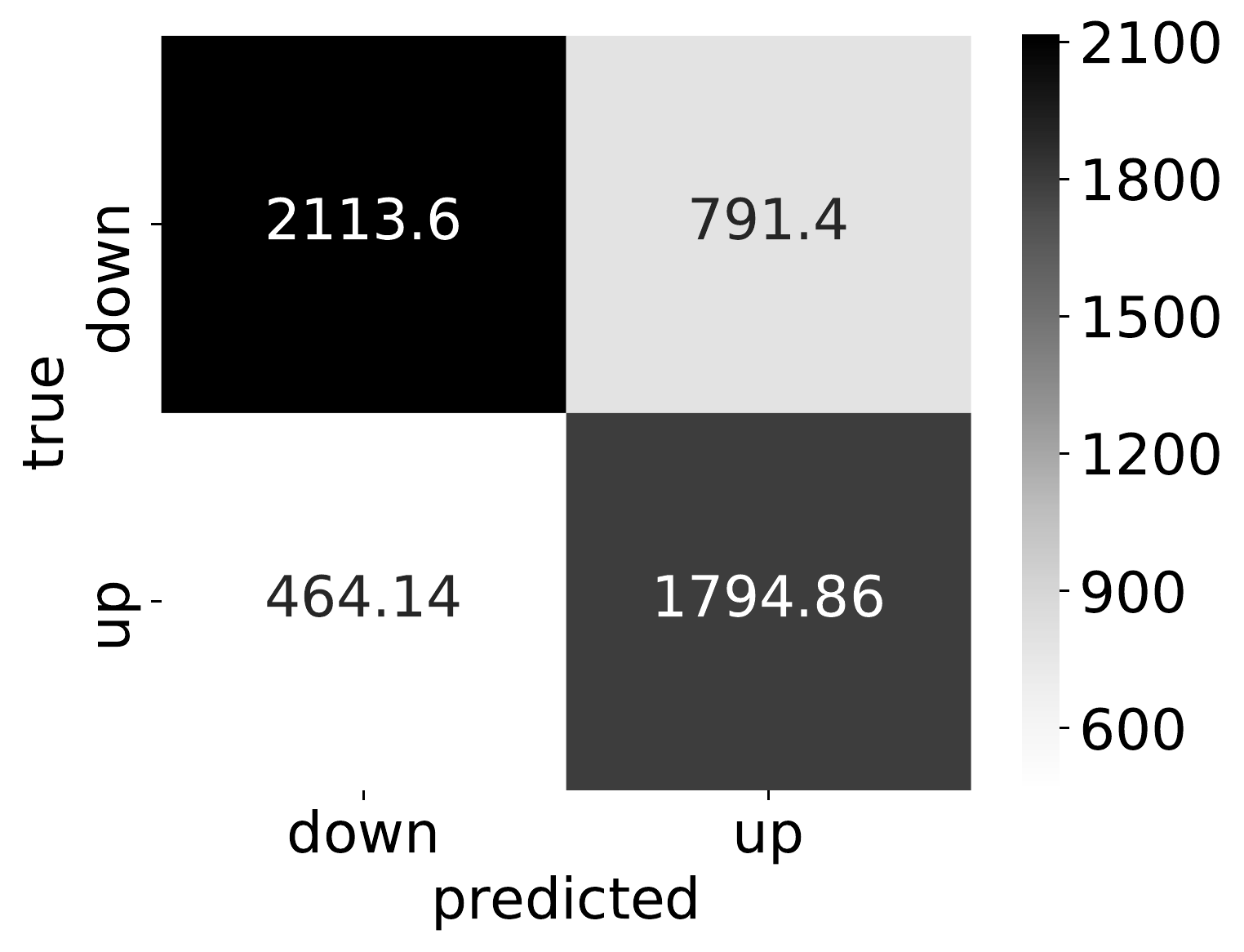}
  \caption{Multi-step, GRU}
  \label{fig:conf_mtx_avg_gru_multi}
\end{subfigure}
\caption{Confusion matrices [individual estimates]}
\label{fig:conf_mtx_avg}
\end{figure}

\begin{figure}
\centering
\begin{subfigure}{.5\textwidth}
  \centering
  \includegraphics[width=.8\linewidth]{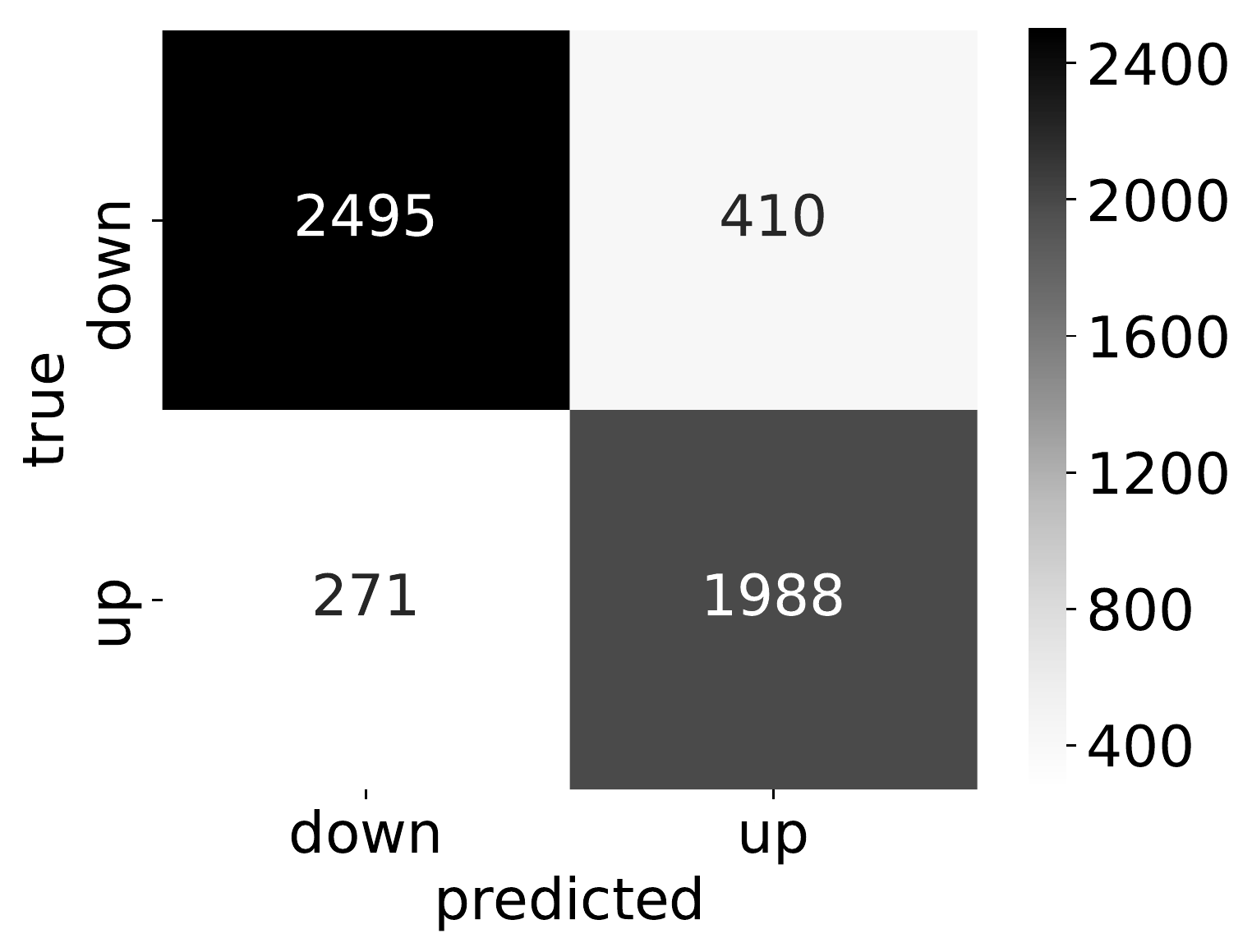}
  \caption{One-step, LSTM}
  \label{fig:conf_mtx_bag_lstm_one}
\end{subfigure}%
\begin{subfigure}{.5\textwidth}
  \centering
  \includegraphics[width=.8\linewidth]{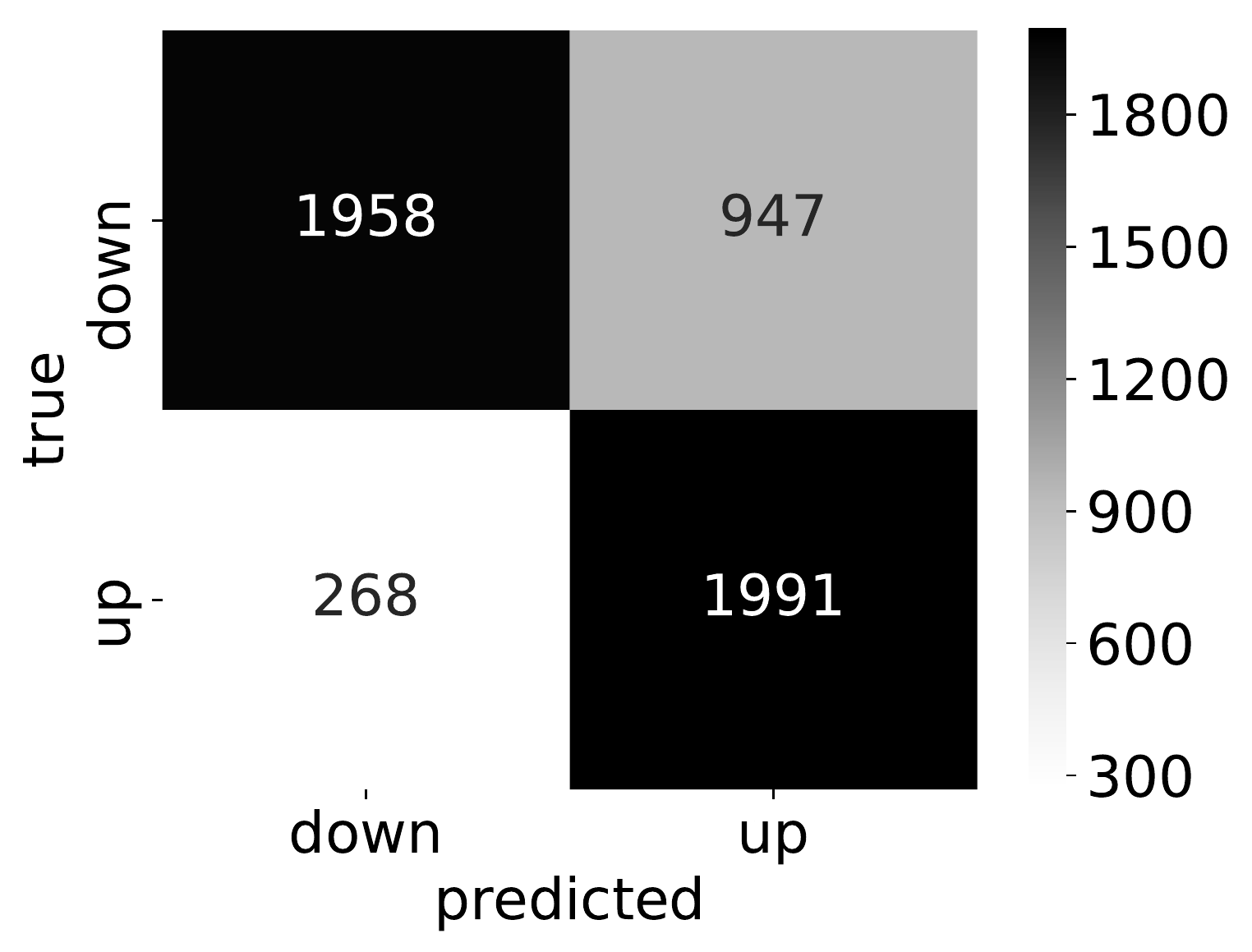}
  \caption{Multi-step, LSTM}
  \label{fig:conf_mtx_bag_lstm_multi}
\end{subfigure}
\begin{subfigure}{.5\textwidth}
  \centering
  \includegraphics[width=.8\linewidth]{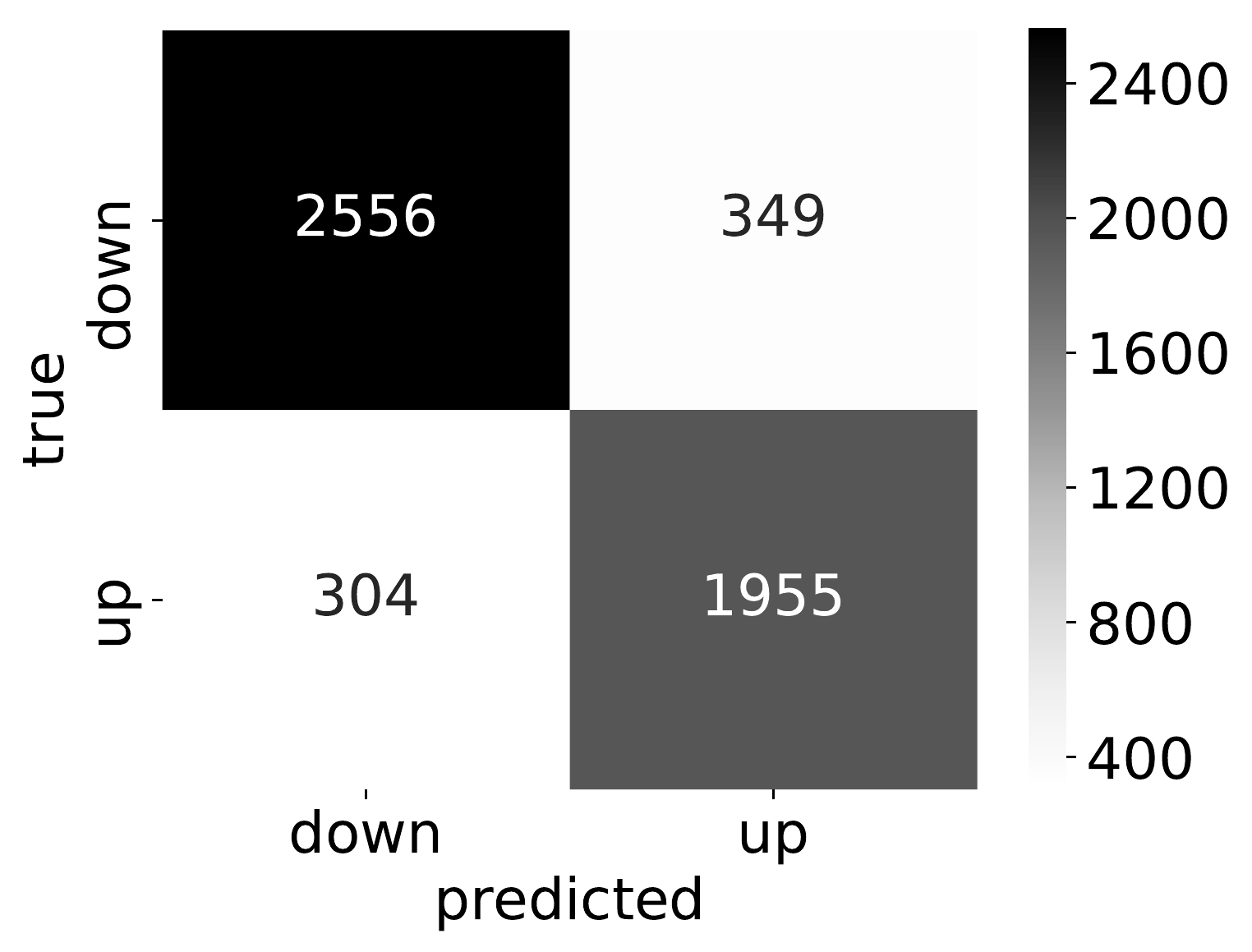}
  \caption{One-step, GRU}
  \label{fig:conf_mtx_bag_gru_one}
\end{subfigure}%
\begin{subfigure}{.5\textwidth}
  \centering
  \includegraphics[width=.8\linewidth]{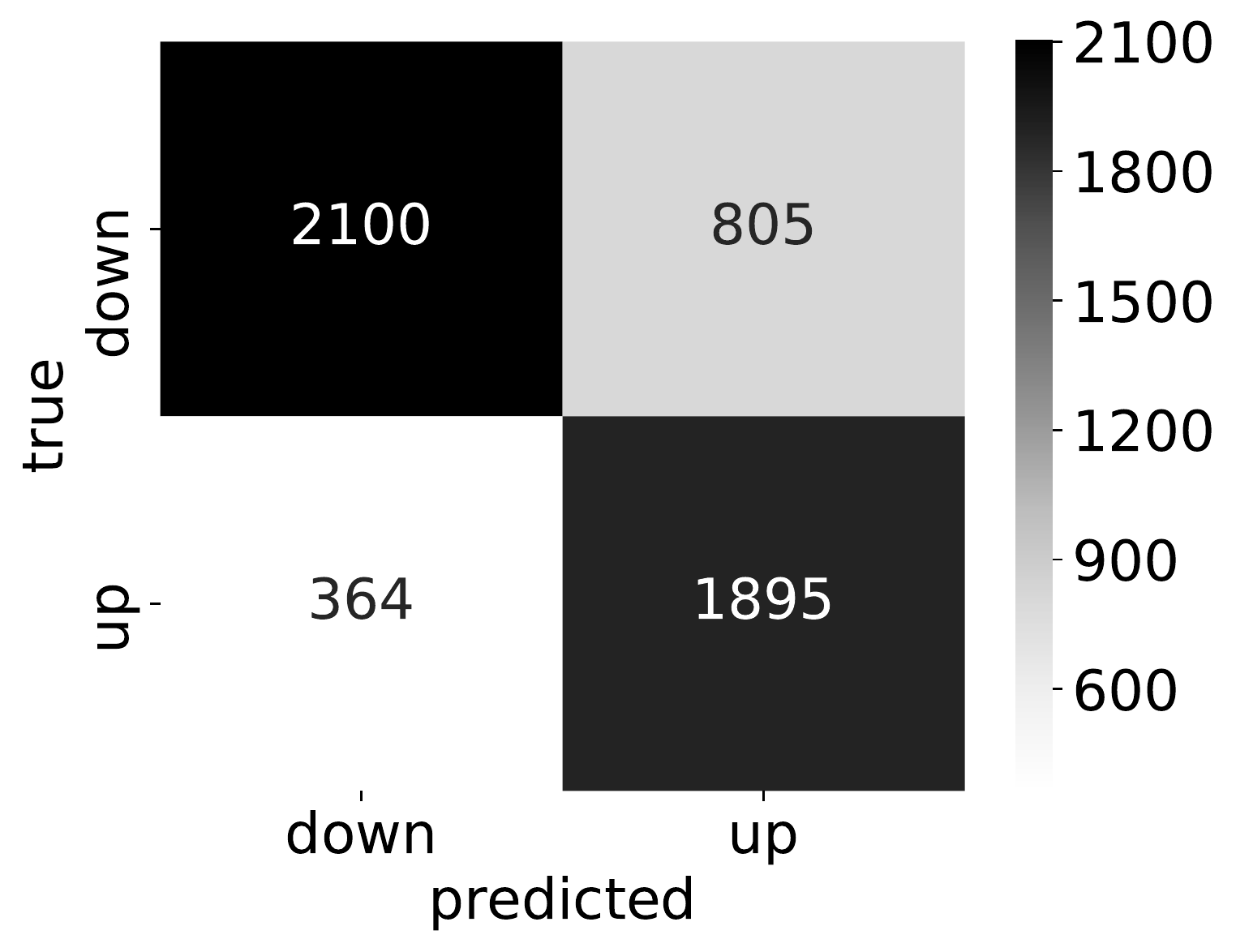}
  \caption{Multi-step, GRU}
  \label{fig:conf_mtx_bag_gru_multi}
\end{subfigure}
\caption{Confusion matrices [bagged estimates]}
\label{fig:conf_mtx_bag}
\end{figure}

Several measures of forecast accuracy and directional accuracy were computed on both evaluation sets.\\
$MSE$, $SMAPE$, $R^2$, $MAE$ and $MedAE$ measure the forecasts' fit, while $accuracy$, $F1$, $precision$ and $recall$ evaluate the quality of the forecasts' changes of direction. Most of these evaluation metrics were computed using scikit-learn (\citet{pedregosa2011scikit}).\\
The results of the regression evaluations are available in Table \ref{table:evaluate_regression_lstm} and Table \ref{table:evaluate_regression_gru}. Table \ref{table:evaluate_classification_lstm} and Table \ref{table:evaluate_classification_gru} show the change-of-direction binary metrics for the test set. Confusion matrices of the directional predictions are displayed in Figure \ref{fig:conf_mtx_avg} and Figure \ref{fig:conf_mtx_bag}.\\
The notations for the tables are the following: b -- bagged estimator; i -- individual estimators; v -- bootstrap left-out validation set; t -- future test set; o -- one-step forecasts; m -- multi-step forecasts.\\
\begin{table}
\centering
\begin{tabular}{ |c|c|c|c|c|c|c|c| }
 \hline
 & $RMSE$ & $MAE$ & $SMAPE$ & $R^2$ & $MedAE$ \\
 \hline
 v-b-o & 31.35 & 20.25 & 25.75 & 0.96 & 12.84 \\
 v-i-o & 33.23 & 21.76 & 28.81 & 0.95 & 13.97 \\
 t-b-o & 45.74 & 30.23 & 22.75 & 0.96 & 19.27 \\
 t-b-m & 90.64 & 62.99 & 41.99 & 0.83 & 39.56 \\
 t-i-o & 49.53 & 33.09 & 25.19 & 0.95 & 21.21 \\
 t-i-m & 103.31 & 71.78 & 44.12 & 0.78 & 46.09 \\
 \hline
\end{tabular}
\caption{Regression evaluation metrics [LSTM]}
\label{table:evaluate_regression_lstm}
\end{table}
\begin{table}
\centering
\begin{tabular}{ |c|c|c|c|c|c|c|c| }
 \hline
 & $RMSE$ & $MAE$ & $SMAPE$ & $R^2$ & $MedAE$ \\
 \hline
 v-b-o & 32.04 & 20.56 & 25.63 & 0.96 & 12.83 \\
 v-i-o & 33.94 & 22.32 & 29.86 & 0.95 & 14.48 \\
 t-b-o & 48.13 & 31.24 & 21.76 & 0.95 & 18.61 \\
 t-b-m & 93.86 & 64.34 & 40.32 & 0.82 & 36.90 \\
 t-i-o & 51.97 & 34.51 & 26.57 & 0.94 & 21.61 \\
 t-i-m & 105.42 & 74.49 & 45.80 & 0.77 & 49.07 \\
 \hline
\end{tabular}
\caption{Regression evaluation metrics [GRU]}
\label{table:evaluate_regression_gru}
\end{table}
\begin{table}
\centering
\begin{tabular}{ |c|c|c|c|c| }
 \hline
 & accuracy & precision & recall & F1 \\
 \hline
 b-o & 0.87 & 0.83 & 0.88 & 0.85 \\
 b-m & 0.76 & 0.68 & 0.88 & 0.77 \\
 i-o & 0.85 & 0.81 & 0.86 & 0.83 \\
 i-m & 0.75 & 0.67 & 0.84 & 0.75 \\
 \hline
\end{tabular}
\caption{Classification evaluation metrics [LSTM]}
\label{table:evaluate_classification_lstm}
\end{table}
\begin{table}
\centering
\begin{tabular}{ |c|c|c|c|c| }
 \hline
 & accuracy & precision & recall & F1 \\
 \hline
 b-o & 0.87 & 0.85 & 0.87 & 0.86 \\
 b-m & 0.77 & 0.70 & 0.84 & 0.76 \\
 i-o & 0.85 & 0.82 & 0.84 & 0.83 \\
 i-m & 0.76 & 0.70 & 0.79 & 0.74 \\
 \hline
\end{tabular}
\caption{Classification evaluation metrics [GRU]}
\label{table:evaluate_classification_gru}
\end{table}
Multi-step forecasts have consistently higher errors than the one-step-ahead predictions. It is not surprising --- just consider the accumulating errors of the iterative forecasting procedure. Lagged values of the target variable proved to be important predictors, especially for forecasting values, rather than just directions. Hence, the accuracy of values that we predict and reuse as inputs, matters a lot. And anyway, the distant future obviously holds much more uncertainty, than the next timestep.\\
All $R^2$ values are close to 0.95 for the single-step predictions, and are around 0.8 for the multi-step predictions. The multi-step $RMSE$ is about twice as large as the one-step. The directional accuracies are roughly 85\% for the one-step, while around 75\% for the multi-step forecasts. For multiple steps, the direction of change forecasts seem to work somewhat better than the value forecasts.\\
We have averaged the resulting estimates from all bootstrap samples, and used the averages as our final predictions --- we used bagging. By comparing the performance of the bagged estimator to the averaged performance of individual estimators, we might evaluate the usefulness of this ensemble method for recurrent neural networks.\\
The bagged estimators produced consistently better results than the individual neural networks. This is true for the regression and the classification problem as well. Thus, it seems that bagging can improve RNN-predictions.\\
The GRU and the LSTM networks are quite close to each other in terms of forecasting performance. The two networks seem to produce very similar forecasts.\\
\citet{papadopoulos2001confidence} found that the bootstrap consistently overestimates prediction interval coverage. It seems to be confirmed by our results, since all of our prediction intervals have a higher coverage than the targeted 90\%. Our residual predictor neural network could probably have been further optimized, in order to generate intervals closer to the desired coverage.\\
CWC's $\mu$ was set to .9, since we aim to generate 90\% prediction intervals. $\eta$ was set to the arbitrary value of 50.\\
The CWC metrics suggest that the prediction intervals of one-step forecasts are better. It is hardly surprising, since the multi-step forecasts reached similar coverage by producing much wider intervals. The coverage of our prediction intervals exceeds the desired level of 90\% in each case, so CWC equals NMPIW. The prediction intervals' evaluation metrics are available in Tables \ref{table:evaluate_pi_lstm} and \ref{table:evaluate_pi_gru}.
%pi metrics
\begin{table}[h!]
\centering
\begin{tabular}{ |c|c|c|c|c| }
 \hline
 & PICP & MPIW & NMPIW & CWC \\
 \hline
 v-o & 0.96 & 99.91 & 0.10 & 0.10 \\
 v-m & 0.98 & 313.68 & 0.33 & 0.33 \\
 t-o & 0.94 & 143.09 & 0.15 & 0.15 \\
 t-m & 0.94 & 371.85 & 0.38 & 0.38 \\
 \hline
\end{tabular}
\caption{PI evaluation metrics [LSTM]}
\label{table:evaluate_pi_lstm}
\end{table}
\begin{table}[h!]
\centering
\begin{tabular}{ |c|c|c|c|c| }
 \hline
 & PICP & MPIW & NMPIW & CWC \\
 \hline
 v-o & 0.96 & 108.11 & 0.11 & 0.11 \\
 v-m & 0.99 & 318.90 & 0.33 & 0.33 \\
 t-o & 0.96 & 161.46 & 0.17 & 0.17 \\
 t-m & 0.95 & 408.34 & 0.42 & 0.42 \\
 \hline
\end{tabular}
\caption{PI evaluation metrics [GRU]}
\label{table:evaluate_pi_gru}
\end{table}
\section{Conclusions and Future Perspectives}
This study aimed to explore and describe several aspects of the application of recurrent neural networks to time series forecasting, though it is by far not comprehensive.\\
Recurrent neural networks are much more flexible and much better suited to time series forecasting than the linear models usually applied. Yet, several practices might help their application, some of which have been presented in this article.\\
We may do time series analysis with the aim of either forecasting future values or understanding the processes driving the time series. Neural networks are particularly bad in the latter. Feature importance measures solve this problem partly. We computed permutation importance scores (mean decrease accuracy). The target variable's lagged values were the most important predictors in our empirical experiment. Seasonality features also seemed important.\\
Another shortcoming of neural networks is the lack of prediction confidence measures. Interval forecasts can be useful for quantifying uncertainty, though producing them for neural networks is nontrivial. Bootstrapping is a simple, yet computationally expensive method that can do it. However, it produced PIs with consistently higher coverage than what we targeted.\\
Multiple-step forecasts generated higher errors than single-step forecasts, as expected. The gaps between the errors seemed smaller in case of direction of change predictions.\\
We found that recurrent neural networks can benefit from bagging.\\
The LSTM and GRU networks showed about the same forecasting performance. It is hard to argue for either of them.\\
This forecasting framework might be enhanced in several ways.\\
During the training process, the states of the LSTM/GRU cells were reset in each batch, so the dependencies between sequences in different batches were not taken into account. We would expect higher accuracies if we took advantage of these relationships in the dataset, though it was difficult with our bootstrapping framework.\\
Our iterative method is probably not the best solution for making multi-step forecasts. A sequence to sequence learning model (e.g., \citet{cho2014learning}, \citet{sutskever2014sequence}) might be a better choice.\\
We could have constructed further input features. Feature engineering is crucial, and there is always room for improvement. Though neural networks are very flexible, so it didn't seem so necessary.\\
Feature importances were only computed for one-step forecasts. It would be worth exploring, if different forecasting horizons require different features to make high quality forecasts. Other measures of variable importance could also be applied.\\
Feature importances are only a tiny step towards understanding recurrent neural networks. The mechanism of RNN cells could and should be explored in much more depth.\\
Bootstrapping is computationally intensive, but with today's ever improving GPUs, it is a feasible algorithm for time series datasets of manageable size. Yet, it is a brute force method, so smarter solutions would be welcome.\\
There are several hyperparameters to optimize --- it is also a disadvantage of neural networks. In this article, we did not aim to find the best parameters. Grid search, or rather random search (\citet{bergstra2012random}) could have helped in finding the ideal settings.\\
The (small) size of real world datasets hinders deep learning methods in the field of time series forecasting. If our variable of interest were only observed quarterly or yearly, we would have to wait several lifetimes to acquire a reasonable amount of data. Even this 2-year hourly bike sharing dataset was way too small to exploit the capabilities of a neural network. It would be very useful if we could train an algorithm on multiple similar datasets and gather some collective knowledge that could be used to make better forecasts for the individual time series. This process of gaining knowledge and applying it to solve different problems is called transfer learning. It is already commonly used in, for example, computer vision (\citet{thrun1996learning}). Transfer learning is most useful when the training data is scarce, so applying it to time series forecasting seems very promising.\\
There is so much left to be done. RNNs clearly deserve a seat in the toolbox of time series forecasting.

%\printbibliography
\bibliography{bib}

\end{document}